\documentclass[11pt,a4paper]{article}
\usepackage[hyperref]{emnlp2020}
\usepackage{times}
\usepackage{latexsym}

\usepackage{microtype}

\aclfinalcopy % Uncomment this line for the final submission

\usepackage{booktabs}
\usepackage{siunitx}
\usepackage{amsmath,amssymb,amsthm}
\usepackage{braket}
\usepackage{enumitem}
\usepackage{stmaryrd}
\usepackage{amsmath}
\usepackage{multirow}
\usepackage{float}
\usepackage{graphicx}
\usepackage[ruled,vlined]{algorithm2e}
\usepackage{ragged2e}

\newcommand\justifiedline[1]{\justifying{\noindent#1}}

\newtheorem{theorem}{Theorem}[section]

\newtheorem{lemma}[theorem]{Lemma}
\newtheorem{property}[theorem]{Property}

\renewcommand\bra[1]{{\langle{#1}|}}
\makeatletter
\renewcommand\ket[1]{%
  \@ifnextchar\bra{\k@t{#1}\!}{\k@t{#1}}%
}
\newcommand\k@t[1]{{|{#1}\rangle}}
\newcommand{\dm}[1] {\llbracket \text{#1} \rrbracket}

\newcommand{\vect}[1] {\overrightarrow{#1}}
\title{Modelling Lexical Ambiguity with Density Matrices}

\author{Francois Meyer \\
  University of Amsterdam \\
  \texttt{francoisrmeyer@gmail.com} \\ \And
  Martha Lewis \\
  University  of Bristol,\\ILLC, University of Amsterdam \\
  \texttt{martha.lewis@bristol.ac.uk} \\}

\date{}

\begin{document}
\maketitle

\begin{abstract}
Words can have multiple senses. Compositional distributional models of meaning have been argued to deal well with finer shades of meaning variation known as polysemy, but are not so well equipped to handle word senses that are etymologically unrelated, or homonymy. Moving from vectors to density matrices allows us to encode a probability distribution over different senses of a word, and can also be accommodated within a compositional distributional model of meaning. In this paper we present three new neural models for learning density matrices from a corpus, and test their ability to discriminate between word senses on a range of compositional datasets. When paired with a particular composition method, our best model outperforms existing vector-based compositional models as well as strong sentence encoders.

\end{abstract}

\section{Introduction}\label{sec:introduction}

An integral part of natural language understanding is the ability to handle lexical ambiguity. Words can have multiple meanings, and the precise meaning of a word only becomes clear when we see it in use - the surrounding context disambiguates it. Word sense disambiguation (WSD) is said to be an `AI complete' problem \cite{navigli2009wsd}, that is, a problem that is at least as hard as any other problem in AI, and as such has been the subject of extensive research. Standard approaches treat WSD as a classification problem: given a word in context, the task is to classify it into one of a range of possible senses \cite{lesk1986automatic,shutze1998wordsense,navigli2009wsd}. A more difficult task is to disambiguate every word in a sentence \cite{chaplot2018knowledge}. A summary of the state of the art is given in \cite{raganato2017word}.
More recently, neural approaches \cite{hadiwinoto2019improved, Huang2019glossbert} use contextualised embeddings as input to WSD systems, together with knowledge from WordNet. Other neural approaches generate multiple sense vectors per word \cite{neelakantan2014mssg, cheng2015syntax} or vectors representing a context \cite{melamud2016context2vec}.

Disambiguation can be costly. Each word should be disambiguated with respect to the correct senses of the other words in the sentence, meaning that the computational complexity of the task can become problematic \cite{chaplot2018knowledge}. Within a compositional framework, the idea is for words to disambiguate automatically in the process of composition \cite{kintsch2001predication, mitchell2008vector, baroni2014frege, boleda2020distributional}.

Within purely vector-based models, the amount of ambiguity that a word vector can represent is limited. \citet{baroni2014frege} argue that distributional vectors work well for polysemy, but less so for homonymy. \citet{piedeleu2015open} extend the vector based model of meaning to encompass homonymy by using the notion of a \emph{density matrix}. These can be used to encode a probability distribution over possible meanings of a word in a single representation. Density matrices can also be accommodated within a compositional framework, allowing the ambiguity encoded in the matrix to be resolved via composition.

We use density matrices within a compositional distributional framework to model word and sentence meaning. We propose three new models for building density matrices, based on neural word embedding models. We survey several composition methods for density matrices and evaluate how well our density matrices encode ambiguity and to what extent the composition methods achieve disambiguation, on four disambiguation datasets that test disambiguation in a compositional setting. One of our models (multi-sense Word2DM) emerges as the best model overall. When paired with a particular composition method (Phaser), multi-sense Word2DM outperforms all other models (including existing baselines and high-performing sentence encoders) on most of the disambiguation tasks. 

\section{Background}\label{sec:background}
Compositional distributional models come in a range of flavours. \citet{mitchell2008vector} use simple element-wise operations on vectors. More recently, neural models of composition \cite{socher2012semantic, bowman2015recursive} and large networks such as BERT \cite{devlin2019bert} have been extremely successful. A third flavour is the type-logical tensor-based models of composition \cite{baroni2010matrices, coecke2010foundations, paperno2014practical,sadrzadeh2018static}.
The tensor-based model of composition works as follows. We choose a vector space $N$ for nouns, and another $S$ for sentences, and represent relational words as multilinear maps over these spaces. Intransitive verbs are represented as linear maps $N \rightarrow S$, i.e. matrices in $N\otimes S$. Transitive verbs are represented as maps from two copies of $N$ to $S$, i.e. order 3 tensors or `cubes' of parameters in $N\otimes S \otimes N$. Composition is performed via \emph{tensor contraction} - an extension of matrix multiplication. Matrices and tensors require many parameters. To alleviate this problem,  \citet{grefenstette2011experimental,grefenstette2011kronecker,kartsaklis2012unified} develop ways of building matrices and tensors from word vectors, some of which are described in section {\ref{cds_models}}.

We use an extension of the tensor-based approach, based on methods given in \citet{piedeleu2015open, bankova2018graded}. Nouns and sentences are represented as \emph{density matrices} and 
relational words (adjectives, verbs, etc.) are represented as \emph{completely positive maps}, which take density matrices to density matrices.

\paragraph{Representing words with density matrices} 
A density matrix over $\mathbb{R}^n$ is a matrix of the form 
\begin{equation} \label{density_matrix}
    \rho = \sum_i p_i \vect{v_i}\vect{v_i}^{\top}
\end{equation}
where $\{p_i\}_i$ are the probabilities assigned to the vectors $\{\vect{v_i}\}_i$. Density matrices over $\mathbb{R}^n$ are:
\begin{enumerate}[noitemsep]
    \item Symmetric:  $\rho^{\top} = \rho$.
    \item Positive semidefinite: $ \forall \vect{x}\! \in\! \mathbb{R}^n,\, \vect{x}^\top\! \rho \vect{x} \geq 0$
    \item Unit trace: $\mathrm{tr}(\rho) = 1$
\end{enumerate}
To represent words, we view each word as a probability distribution over senses, and we view the vectors $\vect{v_i}$ in equation \eqref{density_matrix} as representing its different senses.
For example, the word \emph{bright} could mean \emph{shiny} or \emph{clever}. Suppose that when \emph{bright} is used, it is twice as likely to mean \emph{shiny} as it is to mean \emph{clever}. 
The density matrix for \emph{bright}, denoted $\dm{bright}$, is computed as follows:
\[
\dm{bright} = \frac{2}{3}\vect{\text{shiny}}\,\vect{\text{shiny}}^\top + \frac{1}{3}\vect{\text{clever}}\,\vect{\text{clever}}^\top
\]

\paragraph{Composition with density matrices} 

Since we are working with density matrices, nouns are now maps $N\rightarrow N$, i.e. matrices in $N \otimes N$. Sentences are matrices in $S \otimes S$. This means that intransitive verbs are order-4 tensors that take a noun density matrix as input and give back a sentence density matrix. They live in a space $N \otimes N \otimes S \otimes S$.  Transitive verbs are order-6 tensors. Clearly, these spaces get very big very quickly. 

To deal with this increase in dimensionality, tricks to create completely positive maps out of density matrices have been proposed \cite{lewis2019hyponymy,coecke2020meaning}. This allows composition mechanisms to be specified at the level of density matrices, rather than having to work in the high-order spaces described above. We describe these composition mechanisms in section \ref{dm_composition}.

Other applications of density matrices in NLP include modelling entailment in a compositional setting \citep{balkir2015entailment, bankova2018graded, sadrzadeh2018entailment, lewis2019compositionalhyponymy,bradley2020language}.
 \citet{blacoe2013quantum} also use density matrices to model ambiguity, but in a different setting. 

\paragraph{WSD in compositional distributional semantics}
\citet{baroni2014frege} argue that compositional distributional semantic models are particularly able to pick out the more subtle shades of meaning termed \emph{polysemy}. This idea is used in \cite{mitchell2008vector, grefenstette2011experimental,grefenstette2011kronecker,kartsaklis2013conll}, where a range of semantic composition models are tested on datasets built to distinguish different senses of words in context. Neural and distributional models for disambiguation are compared in \citet{milajevs2014evaluating}, and the role of ellipsis in disambiguation is investigated in \citet{wijnholds2019ellipsis}.

\section{Methods}\label{sec:methods}

\subsection{Density Matrix Models}

We now introduce the three novel methods that we propose for building density matrices.

\subsubsection{BERT2DM}
\SetAlgoLined
\begin{algorithm}
    \For{each sentence $s$ in a corpus}{
    	 \justifiedline{Process $s$ with BERT.}
    	
    	 \justifiedline{Extract and store the contextualised embeddings produced by BERT for the words in $s$.}
    }   
    
    {\justifiedline{Discard all contextualised embeddings corresponding to stop words.}}
    
    {\justifiedline{Apply PCA/SVD to the remaining contextualised embeddings.} }
    
    \For{each word $v$ in the vocabulary}{
        \justifiedline{Compute the density matrix of $v$ as}
        \[
            \llbracket v \rrbracket = \sum_{i \in \mathrm{ind}(v)} \vect{{v_i}}\,\vect{{v_i}}^\top,
        \]
        where $\mathrm{ind}(v)$ are the indices at which\\ the word $v$ occurs in the corpus and $\vect{v_i}$\\ is the reduced embedding for $v$.}
 \caption{BERT2DM training}
 \label{bert2dm_alg}
\end{algorithm}

BERT \cite{devlin2019bert} produces \emph{contextualised embeddings} for words and sentences. Given a sentence, it produces vectors for each word that are specific to that particular context (BERT actually models subword units, but we average the subword embeddings of a word to obtain a contextualised word embedding). BERT2DM uses the contextualised embeddings of BERT to build density matrices that encode multiple senses of a word. BERT is applied to a corpus and the contextualised embeddings for a word $w$ are combined to compute $w$'s density matrix according to equation \eqref{density_matrix}. The procedure is outlined in algorithm \ref{bert2dm_alg}.

Since the vectors produced by BERT are fairly large, we apply a dimensionality reduction step (either PCA or SVD) over all content word embeddings before combining to form a density matrix.

We also experiment with clustering the contextual embeddings of a word and applying dimensionality reduction to the cluster centroids instead of the contextualised embeddings. The motivation for this is that clustering contextualised embeddings can produce clusters that correspond to distinct senses (as shown by \citet{wiedemann2019doesbertmakeanysense}).

\subsubsection{Word2DM}

\begin{algorithm}
\SetAlgoLined
    \For{each word $v$ in the vocabulary}{
    	 \justifiedline{Randomly initialise a $n\times m$ matrix $B_v$.}
    } 
    \For{each target word $w_t$ in the corpus}{
    	\For{each context word $w_c$}{
        	 \justifiedline{Sample $K$ negative samples from some noise distribution.}
        	 
             \justifiedline{Maximise equation \ref{word2dm_negative_sampling} with respect to $B_t$, $B_c$, and $B_{w_k}$ for $k=1, ... K$.} 
        } 
    } 
    \For{each word $v$ in the vocabulary}{
    	 \justifiedline{Compute  its density matrix as $A_v = B_v B_v^\top $.}
    } 
 \caption{Word2DM training}
 \label{word2dm_alg}
\end{algorithm}

Word2DM is an extension of Word2Vec \citep{mikolov2013word2vec1, mikolov2013word2vec2} skip-gram with negative sampling (SGNS). SGNS modifies word vectors to become closer to words they do occur with, and further away from words they don't occur with (the negative samples). When extending the SGNS algorithm to produce density matrices, we must ensure
that the matrices satisfy the conditions resulting from their definition: symmetry, positive semidefiniteness, and unit trace. The first and last are easy to enforce, but preserving positivity is more challenging. To preserve positivity, 
we utilise the following property of positive semi-definiteness:
\begin{property}\label{property_psd2}
For \emph{any} matrix $B$, the product $BB^{\top}$ is positive semi-definite.
\end{property}

We enforce positive semi-definiteness
by training the weights of an intermediary matrix $B$ and computing our density matrix as $A = BB^{\top}$.
By updating the weights of $B$ and computing $A$ we indirectly train positive semi-definite matrices. 
We modify the training objective of SGNS to maximise the similarity of the density matrices of co-occurring words.
The objective function at each target-context prediction is then:
\begin{equation} \label{word2dm_negative_sampling}
    J(\theta) = \log \sigma(\mathrm{tr}(A_t A_c)) + \sum_{k=1}^K  \log \sigma(-\mathrm{tr}(A_t A_{w_k})) 
\end{equation} 
where $A_t$ and $A_c$ are the the density matrices of the target and context words respectively, $A_1, A_2, ..., A_K$ are the density matrices of $K$ negative samples, and $\theta$ is the set of weights of the intermediary matrices $B_t, B_c$ and $B_1, B_2, ..., B_K$. 

Word2DM is a straightforward extension of Word2Vec for learning density matrices. However, it turns out that enforcing positive semi-definiteness by introducing intermediary matrices leads to suboptimal training updates. This can be shown by examining the gradients of equation \eqref{word2dm_negative_sampling} with respect to the intermediary matrices (derivation in supplementary material). 

\subsubsection{Multi-sense Word2DM}
Multi-sense Word2DM is a modification of Word2DM designed to overcome the gradient issues of Word2DM and to explicitly model ambiguity. Multi-sense Word2DM achieves this through the following changes to Word2DM:
\begin{itemize}[noitemsep,leftmargin=*]
    \item The columns of the intermediary $n \times m$ matrix $B$ now represent the $m$ different senses of the word. Each sense of a word has its own $n$-dimensional embedding. The density matrix of a word is still computed as before, and can be expressed in terms of the sense embeddings as
    \begin{equation}\label{ms_columns}
        A = BB^\top = \sum_{i=1}^{m} \vect{b_i}\,\vect{b_i}^\top
    \end{equation}
    where $\vect{b_1}, ..., \vect{b_m}$ are the columns of $B$ corresponding to different senses. 
    \item Each word is also associated with a single vector $v_w$, which represents it as a context word.
    \item 
    The following objective function is maximised: 
    \begin{equation}\label{ms_word2dm_negative_sampling}
    J(\theta) = \log \sigma({b_t}^\top\! c_t) + \sum_{k=1}^K  \log \sigma(-{b_t}^\top\! v_{w_k})
    \end{equation}
    where $c_t$ is the sum of context vectors for all words surrounding the target word and $b_t$ is the the embedding for the relevant sense of the target word. We select $b_t$ by finding the column of $B_t$ most similar to $c_t$ (measured by either cosine similarity or dot product). The full training procedure is outlined in algorithm \ref{ms_word2dm_alg}.
\end{itemize}

Multi-sense Word2DM explicitly models ambiguity by letting the columns of the intermediary matrix represent the different senses of a word. During training the column closest to the context embedding is selected as the relevant sense embedding and only this column is updated. This enables the model to avoid the gradient issues of Word2DM. The objective function being maximised (equation \ref{ms_word2dm_negative_sampling}) has the same gradient as Word2Vec and therefore does not lead to suboptimal training updates.

\subsection{Composition methods}
\label{dm_composition}
The composition methods we use are based on methods in \cite{lewis2019compositionalhyponymy,coecke2020meaning}. These reduce the high-dimensional representations needed for relation words to composition of density matrices. The relational word is seen as a map that takes nouns as arguments.
The composition methods are as follows, using the example of an adjective modifying a noun:
\begin{description}[noitemsep]
    \item[Add]: $\dm{adj} + \dm{noun}$
    \item[Mult]: $\dm{adj} \odot \dm{noun}$
    \item[Tensor]: $\dm{adj} \otimes \dm{adj} \times \dm{noun}$, where $\otimes$ denotes the Kronecker product and $\times$ denotes tensor contraction.
    \item[Phaser]: $\dm{adj}^{1/2}\dm{noun}\dm{adj}^{1/2}$
\end{description}
More complex phrases are combined according to their parse. So, a transitive sentence modified with an adjective is composed as $(\text{subj} (\text{verb}(\text{adj obj})))$. For example, composing the sentence \emph{Bob likes old cars} would consist of the following steps:

{\centering
 $\dm{old cars} = f(\dm{old}, \dm{cars})$

 $\dm{likes old cars} = f(\dm{likes}, \dm{old cars})$

$\dm{Bob likes old cars} = f(\dm{Bob}, \dm{likes old cars}) $}

\noindent where the composer $f$ can be substituted by any of the composition methods listed above.

\begin{algorithm}[t]
\SetAlgoLined
    \For{each word $w$ in the vocabulary}{
    	 \justifiedline{Randomly initialise a $n\times m$ matrix $B_w$ and a $n$-dimensional vector $v_w$.}
    } 
    \For{each target word $w_t$ in the corpus}{
        \justifiedline{Sum the context vectors of the words surrounding $w_t$ within a window of size $2l$ to get a context embedding $c_t$:}
            \[c_t = \sum_{i=t-l, i\neq t}^{t+l} v_i\]
            
        \justifiedline{Compute the similarity (with either cosine similarity or dot product) of the columns $b_1, ..., b_m$ of $B_t$ and $c_t$ and extract the most similar column as $b_t$ (the embedding of the relevant sense).}
        
        \justifiedline{Sample $K$ negative samples from some noise distribution.
        Maximise equation \ref{ms_word2dm_negative_sampling} with regards to $b_t$, $c_t$, and $v_{w_k}$ for $k=1, ... K$.}
    } 
    \For{each word $v$ in the vocabulary}{
    	\justifiedline{Compute its density matrix as $A_v = B_v B_v^\top $.}
    } 
 \caption{MS-Word2DM training}
 \label{ms_word2dm_alg}
\end{algorithm}

\section{Experimental Setup}\label{sec:experiments}

We evaluate our models on three tasks - word similarity, disambiguation, and a word level ambiguity analysis. The code for training and evaluating our models has been made available at \texttt{https://github.com/francois-meyer/ lexical-ambiguity-dms}.
In this section we introduce the experimental setup used in all of these tasks.

\subsection{Baselines}

Throughout the experiments we compare our models to existing word and sentence embedding models. For our word embedding baselines we use embeddings produced by three existing models - Word2Vec, GloVe, and FastText. We use the publicly available\footnote{https://github.com/gijswijnholds/compdisteval-ellipsis} embeddings trained by \citet{wijnholds2019ellipsis}. The embeddings are 300-dimensional and were trained on the combined and lemmatised ukWaC and Wackypedia corpora\footnote{wacky.sslmit.unibo.it}.
In the tasks that involve sentence-level semantics (the disambiguation tasks) we compare our models to existing compositional distributional semantic (CDS) models and neural sentence encoders.

\subsubsection{CDS models}
\label{cds_models}
CDS models compute a sentence vector as a function of the distributional vectors of the words in the sentence. We use the pre-trained word embeddings of \citet{wijnholds2019ellipsis} and compute sentence embeddings by either summing, element-wise multiplying, or applying tensor-based composition. 
For the phrase \emph{big house} with vectors $\overrightarrow{\text{big}}$ and $\overrightarrow{\text{house}}$ the different compositional distributional methods will be computed as follows:
\begin{description}[noitemsep]
    \item[Add]: $\overrightarrow{\text{big}} + \overrightarrow{\text{house}}$
    \item[Mult]: $\overrightarrow{\text{big}} \odot \overrightarrow{\text{house}}$
    \item[Tensor]: $\overrightarrow{\text{big}} \times \overrightarrow{\text{big}}^\top \times \,\,\, \overrightarrow{\text{house}}$
\end{description}

\subsubsection{Sentence encoders}
We compare our models to two well-known neural sentence encoders - InferSent \citep{conneau2017infersent} and BERT \citep{devlin2019bert}. 
InferSent embeddings are 4096-dimensional, which is much larger than the word embeddings used in our CDS baselines. We use two pre-trained InferSent models that are publicly available\footnote{https://github.com/facebookresearch/InferSent}, referred to as InferSent1 and InferSent2 in our results. 
We also compare our models to BERT as a sentence encoder. 
BERT produces an embedding for the entire sentence by adding a special classification token (\verb|[CLS]|) to the start of every sequence. 
When BERT is used in a sentence-level task, the \verb|[CLS]| embedding can be used as a semantic representation for the entire sentence.

Some of our evaluation data contains phrases that are not fully formed sentences. To ensure a fair comparison, we convert all phrases to fully formed sentences for evaluation of the sentence encoders. We added ``the'' before noun phrases and converted verbs to their present tense form.

\subsubsection{Context2DM}

We also compare our models to a baseline density matrix model, which we call Context2DM. It is based on the procedure of \citet{shutze1998wordsense} for building multi-sense embeddings. 
Context2DM builds the density matrix of a word $w$ as follows:
\begin{enumerate}[noitemsep,leftmargin=*]
    \item Context embeddings are obtained for all the contexts in which $w$ occurs (computed by summing the pre-trained embeddings of all the words that occur around $w$ in a particular context).
    \item These context embeddings are clustered  (using hierarchical agglomerative clustering for $k=2,...,10$ and the variance reduction criterion to select the number of clusters) and the resulting cluster centroids subsequently represent the different senses of $w$.
    \item The density matrix of $w$ is computed as the mixture of its sense embeddings i.e. the sum of the outer products of the cluster centroids, normalised to have unit trace. 
\end{enumerate}
For the pre-trained word embeddings required for step 1 of the above procedure, we use 17-dimensional word embeddings, trained with the gensim implementation\footnote{https://radimrehurek.com/gensim/models/word2vec} of Word2Vec on the combined ukWaC+Wackypedia corpus.

\begin{figure*}[t]\footnotesize
    \centering
	\begin{tabular}{llllcc} 
		\toprule
	     Data set  & Format & High similarity example  & Low similarity example & \# Pairs & \# Annotators\\
	     \midrule
	     \multirow{2}{*}{ML2008}  & \multirow{2}{*}{S\underline{V}} & value slump & value slump & \multirow{2}{*}{120} & \multirow{2}{*}{53}\\
	     & & value decline & value slouch & & \\
	     \midrule
	     \multirow{2}{*}{GS2011} & \multirow{2}{*}{S\underline{V}O} & people buy house & people buy house & \multirow{2}{*}{200} & \multirow{2}{*}{25}\\
	     & & people purchase house & people bribe house & &\\
	     \midrule
	     \multirow{2}{*}{GS2012} & \multirow{2}{*}{AS\underline{V}AO} & local family run small hotel & local family run small hotel & \multirow{2}{*}{194} & \multirow{2}{*}{50}\\
	     & & local family operate small hotel & local family move small hotel & &\\
	     \midrule
	     KS2013 & \multirow{2}{*}{AS\underline{V}AO} & young woman file long nail & young woman file long nail & \multirow{2}{*}{194} & \multirow{2}{*}{43}\\
	     -CoNLL& & young woman smooth long nail & young woman register long nail & &\\
		\bottomrule
	\end{tabular}
	\captionof{table}{Details of the disambiguation data sets (A: adjective, S: subject, V: verb, O: object).}\label{dataset_details}
\end{figure*}

\subsection{Training} \label{subsec:training}

All our density matrices are $17 \times 17$ (so 289 parameters). 
This is closest in size to the 300-dimensional baseline embeddings.
We train our Word2DM and multi-sense Word2DM models on the ukWaC+Wackypedia corpus, consisting of 2.8 billion words. Hyperparameters are provided in the supplementary material. Training these models on a single GPU (Nvidia GeForce GTX 1080 Ti) with 60GB of memory takes around 20 hours per iteration of the training corpus.
We present results for four different multi-sense Word2DM models. Two use cosine similarity to compare sense vectors to context vectors, while the other two use the dot product. We also vary the number of senses modelled (the number of columns in the intermediary matrix) between 5 and 10.

We present results for four different BERT2DM models. Two of these cluster the BERT representations into senses before dimensionality reduction, while the other two do not. One of the advantages of clustering the representations is that it reduces the size of the matrix on which dimensionality reduction is applied, so it becomes computationally feasible to train on a larger corpus. We train the unclustered variants on a 10-million word subcorpus of Wackypedia, and the clustered variants on a 20-million word subcorpus. We also vary the dimensionality reduction algorithm between PCA and SVD, to test whether or not centering the contextual embeddings before dimensionality reduction makes any difference. Training BERT2DM takes only a few hours on a 16-core CPU (Intel Xeon Gold 6130) but requires around 4.5GB of memory per 1 million words that it is trained on.

\subsection{Data Sets}
We test our models on data sets designed to test disambiguation in a compositional setting.
Data sets for this task contain sentence pairs with:
\begin{itemize}[noitemsep,topsep=0pt]
    \item An ambiguous \emph{target} word used in a disambiguating phrase.
    \item A \emph{landmark} word that has the same meaning as one of the target word's senses.
    \item Human judgements of how similar the meaning of the phrase is when the ambiguous word is replaced by the landmark word.
\end{itemize}
We use four disambiguation data sets to evaluate our models. Three of the four data sets - GS2011 \citep{grefenstette2011experimental}, GS2012, and KS2013-CoNLL \citep{kartsaklis2013conll} - are publicly available\footnote{http://compling.eecs.qmul.ac.uk/resources/}, while ML2008 \citep{mitchell2008vector} was obtained privately from the authors of \citet{wijnholds2019ellipsis}. 
We show examples and statistics of the data sets in table \ref{dataset_details}.

\section{Results}

We introduce each of the evaluation tasks and present our results. For multi-sense Word2DM and BERT2DM we trained four models each, with different hyperparameter settings (as described in section \ref{subsec:training} and listed in table \ref{wordsim_results}). 

\begin{figure}[h]\small
    \centering
	\begin{tabular}{lccccc} 
		\toprule
	     & \textbf{RG} & \textbf{WS} & \textbf{MC} & \textbf{SL} & \textbf{MEN} \\
	     \midrule
	     Word2Vec & .818 &	\textbf{.662}	& \textbf{.765} &	.404&	\textbf{.781} \\
	     GloVe & \textbf{.826} & .571 &	.732 &	.399 &	.773\\
	     FastText & .767 &	.517 &	.682 &	\textbf{.404} &	.768 \\
	     \midrule
	     Context2DM & .228 & .234 & .331	& .094 & .267\\
	     Word2DM  & .541 & .473 & .452	& .157 & .540 \\
	     \midrule 
         MS-Word2DM & & & & & \\
         - cos, 5 senses & \textbf{.768} & .556 & \textbf{.670} & \textbf{.290} & .680 \\
         - cos, 10 senses & .727	& .580 & .659 & .256 & \textbf{.682} \\
         - dot, 5 senses & .662	& .578 &	.568 & .247 & .663\\
         - dot, 10 senses & .679 & \textbf{.596} & .612 & .281 & .663\\
         \midrule 
         BERT2DM & & & & & \\
         - PCA & .452 & .275 & .388 & .226 & .351 \\
         - SVD & .428 & .317 & .392 & .234 & .327 \\
         - PCA + cluster & .383	& .219 & .381 & .153 & .251\\
         - SVD + cluster & .315	& .205	&	.294 & .091 & .207\\
		\bottomrule
	\end{tabular}
	\captionof{table}{Spearman $\rho$ obtained on word similarity tasks.} 	\label{wordsim_results}
	%\vspace{-2cm}
\end{figure}

\subsection{Word Similarity}

To validate the quality of our density matrices as general semantic representations we evaluate them on
 the following standard word similarity data sets: RG \citep{rubenstein1965rg}, WS \citep{finkelstein2001ws353}, MC \citep{miller1991mc}, SL \citep{hill2015simlex999}, and MEN \citep{bruni2012men}. We use the evaluation scripts and data sets made publicly  available\footnote{https://github.com/mfaruqui/eval-word-vectors} by \citet{faruqui2014eval}.
The results are shown in table \ref{wordsim_results}. 

\newcommand\x{0.042}
\begin{figure}[t]		
    \small
	\begin{tabular}{@{}p{0.14\textwidth}p{\x\textwidth}p{\x\textwidth}p{\x\textwidth}p{\x\textwidth}p{\x\textwidth}}
		\toprule
	     & Verb & Mult & Add & Tensor & Phaser \\
		\toprule	
		Word2Vec & .215 & .256 & .299 & .231 \\
        GloVe & .332 & .098 & .304 & \textbf{.397} \\
        FastText & .181 & .281 & .198 & .137 \\
        \midrule
        BERT & \multicolumn{5}{c}{.140} \\
        InferSent1 & \multicolumn{5}{c}{.207} \\
        InferSent2 & \multicolumn{5}{c}{.174} \\
        \midrule
        Context2DM & -.069 & -.025 & -.058 & -.025 & -.064 \\
        Word2DM & -.022 &\phantom{-}.057 & \phantom{-}.010 & \phantom{-}.057 & -.007 \\
        \midrule
        MS-Word2DM &&&&&\\
        - cos, 5 senses & \phantom{-}.235 & \phantom{-}.195 & \phantom{-}.254 & \phantom{-}.195 & \textbf{\phantom{-}.328} \\
        - cos, 10 senses & \phantom{-}.248 & \phantom{-}.204 & \phantom{-}.210 & \phantom{-}.204 & \phantom{-}.217 \\
        - dot, 5 senses & \phantom{-}.216 & \phantom{-}.145 & \phantom{-}.280 & \phantom{-}.145 & \phantom{-}.311 \\
        - dot, 10 senses & \phantom{-}.157 & \phantom{-}.195 & \phantom{-}.170 & \phantom{-}.195 & \phantom{-}.325 \\
        \midrule
        BERT2DM &&&&&\\
        - PCA & -.055 & -.016 & -.101 & -.016 & -.114 \\
        - SVD & \phantom{-}.072 & \phantom{-}.170 & \phantom{-}.075 & \phantom{-}.170 & \phantom{-}.067 \\
        - PCA + cluster & \phantom{-}.105 & \phantom{-}.170 & \phantom{-}.062 & \phantom{-}.170 & \phantom{-}.130 \\
        - SVD + cluster & \phantom{-}.090 & -.002 & \phantom{-}.057 & -.002 & \phantom{-}.049 \\
		\bottomrule
	\end{tabular}
	\captionof{table}{Spearman $\rho$ obtained on ML2008. } \label{ml2008_table}
\end{figure}

\begin{figure}[t]		
    \small
	\begin{tabular}{@{}p{0.14\textwidth}p{\x\textwidth}p{\x\textwidth}p{\x\textwidth}p{\x\textwidth}p{\x\textwidth}}
		\toprule
	     & Verb & Mult & Add & Tensor & Phaser \\
		\toprule	
		Word2Vec & .209 & .203 & .268 & .204 \\
        GloVe & \textbf{.304} & .211 & .252 & .256 \\
        FastText & .210 & .187 & .154 & .185 \\
        \midrule
        BERT & \multicolumn{5}{c}{.266} \\
        InferSent1 & \multicolumn{5}{c}{.241} \\
        InferSent2 & \multicolumn{5}{c}{.194} \\
        \midrule
        Context2DM & \phantom{-}.037 & -.027 & \phantom{-}.012 & -.021 & \phantom{-}.036 \\
        Word2DM & -.059 & \phantom{-}.001 & \phantom{-}.019 & -.064 & -.039 \\
        \midrule
        MS-Word2DM &&&&&\\
        - cos, 5 senses & \phantom{-}.187 & \phantom{-}.286 & \phantom{-}.206 & \phantom{-}.289 & \phantom{-}\underline{\textbf{.365}} \\
        - cos, 10 senses & \phantom{-}.091 & \phantom{-}.237 & \phantom{-}.161 & \phantom{-}.200 & \phantom{-}\underline{.323} \\
        - dot, 5 senses & -.016 & \phantom{-}.091 & -.002 & \phantom{-}.010 & \phantom{-}.077 \\
        - dot, 10 senses & -.021 & \phantom{-}.116 & \phantom{-}.025 & \phantom{-}.131 & \phantom{-}.118 \\
        \midrule
        BERT2DM &&&&&\\
        PCA& -.024 & -.107 & -.001 & -.097 & -.046 \\
        SVD & -.105 & -.304 & -.078 & -.057 & -.074\\
        PCA + cluster & -.056 & \phantom{-}.030 & \phantom{-}.008 & \phantom{-}.007 & -.028 \\
        SVD + cluster & \phantom{-}.045 & \phantom{-}.013 & \phantom{-}.031 & \phantom{-}.037 & -.029 \\
		\bottomrule
	\end{tabular}
	\captionof{table}{Spearman $\rho$ obtained on GS2011. } 
	\label{gs2011_table}
\end{figure}

Multi-sense Word2DM performs best out of all the density matrix models, achieving scores comparable to the word embeddings. It substantially improves upon Word2DM, supporting our theoretical findings about Word2DM's learning issues. Using cosine similarity to select the relevant sense results in slightly better scores. The BERT2DM models perform worst of all our models, but still demonstrate some ability to judge word similarity. There is no clear performance difference between using PCA or SVD for dimensionality reduction. Clustering the BERT representations before dimensionality reduction leads to worse correlation scores.

\subsection{Disambiguation} \label{subsec:disambiguation_results}

The results we obtain on the disambiguation data sets are presented in tables \ref{ml2008_table} to \ref{ks2013conll_table}. 
In each of these tables our density matrix models are compared to our baselines.
Column headings specify the composition methods used to compute the phrase representation. These do not apply to the sentence encoders (BERT and InferSent).
The rightmost composition method (Phaser) does not apply to the CDS models.
The leftmost column (Verb) compares the semantic representations of the verbs without composition.
The best performing models, among the baselines and the density matrices, are indicated in bold.
We compare the best-performing density matrix models to the best-performing baseline using a one-sided paired t-test (applying the Bonferroni correction to account for multiple comparisons). We indicate statistically significant improvements over the baseline models, or statistically equivalent scores, by underlining the corresponding scores.

Multi-sense Word2DM is by far the best performing density matrix model. It outperforms all the baseline models on 3 out of the 4 data sets.
Among all the composition methods, Phaser most consistently achieves high correlation scores (especially on the more complex data sets). 
In some cases BERT2DM achieves correlation scores that are comparable to multi-sense Word2DM and the baselines. But in general the BERT2DM density matrices cannot reliably be used to achieve disambiguation.

\begin{figure}[t]		
    \small
	\begin{tabular}{@{}p{0.14\textwidth}p{\x\textwidth}p{\x\textwidth}p{\x\textwidth}p{\x\textwidth}p{\x\textwidth}}
		\toprule
	     & Verb & Mult & Add & Tensor & Phaser \\
		\toprule	
		Word2Vec & .270 & .155 & .334 & .260 \\
        GloVe & .413 & .219 & .297 & .231 \\
        FastText & .302 & .176 & .175 & .264 \\
        \midrule
        BERT & \multicolumn{5}{c}{ \textbf{.471}} \\
        InferSent1 & \multicolumn{5}{c}{.370} \\
        InferSent2 & \multicolumn{5}{c}{.372} \\
        \midrule
        Context2DM & -.025 & \phantom{-}.015 & -.063 & -.033 & \phantom{-}.005 \\
        Word2DM & -.047 & \phantom{-}.019 & -.092 & \phantom{-}.043 & \phantom{-}.074 \\
        \midrule
        MS-Word2DM &&&&&\\
        - cos, 5 senses & \phantom{-}.266 & \phantom{-}.203 & \phantom{-}.270 & \phantom{-}.329 & \phantom{-}\underline{\textbf{.500}} \\
        - cos, 10 senses & \phantom{-}.214 & \phantom{-}.208 & \phantom{-}.263 & \phantom{-}.304 & \phantom{-}.397 \\
        - dot, 5 senses & -.067 & \phantom{-}.068 & -.103 & \phantom{-}.022 & \phantom{-}.126 \\
        - dot, 10 senses & -.059 & \phantom{-}.082 & -.040 & \phantom{-}.068 & \phantom{-}.126 \\
        \midrule
        BERT2DM &&&&&\\
        - PCA& \phantom{-}.025 & \phantom{-}.031 & \phantom{-}.122 & \phantom{-}.056 & \phantom{-}.056 \\
        - SVD & -.071 & \phantom{-}.001 & -.042 & -.037 & -.056 \\
        - PCA + cluster & -.232 & -.073 & -.141 & -.155 & -.232 \\
        - SVD + cluster & -.132 & -.117 & -.154 & -.089 & -.172\\
		\bottomrule
	\end{tabular}
	\captionof{table}{Spearman $\rho$ obtained on GS2012. } 
	\label{gs2012_table}
\end{figure}

\begin{figure}[t]		
    \small
	\begin{tabular}{@{}p{0.14\textwidth}p{\x\textwidth}p{\x\textwidth}p{\x\textwidth}p{\x\textwidth}p{\x\textwidth}}
		\toprule
	     & Verb & Mult & Add & Tensor & Phaser \\
		\toprule	
		Word2Vec & .201 & .222 & .194 & .190 \\
        GloVe & .152 & .154 & .127 & .083 \\
        FastText & .081 & .285 & .073 & .196 \\
        \midrule
        BERT & \multicolumn{5}{c}{ \textbf{.314}} \\
        InferSent1 & \multicolumn{5}{c}{.187} \\
        InferSent2 & \multicolumn{5}{c}{.190} \\
        \midrule
        Context2DM & -.017 & -.074 & -.037 & -.064 & -.006 \\
        Word2DM & \phantom{-}.149 & -.118 & \phantom{-}.114 & -.014 & \phantom{-}.081 \\
        \midrule
        MS-Word2DM &&&&&\\
        - cos, 5 senses & \phantom{-}.135 & \phantom{-}.075 & \phantom{-}.190 & \phantom{-}.147 & \phantom{-}\underline{.309} \\
        - cos, 10 senses & \phantom{-}.171 & \phantom{-}.008 & \phantom{-}.207 & \phantom{-}.118 & \phantom{-}.288 \\
        - dot, 5 senses & -.026 & \phantom{-}.034 & \phantom{-}.039 & \phantom{-}.117 & \phantom{-}.275 \\
        - dot, 10 senses & \phantom{-}.094 & \phantom{-}.016 & \phantom{-}.070 & \phantom{-}.053 & \phantom{-}\underline{\textbf{.345}} \\
         \midrule
        BERT2DM &&&&&\\
        - PCA & \phantom{-}.084 & \phantom{-}.082 & \phantom{-}.089 & \phantom{-}.081 & \phantom{-}.187 \\
        - SVD & \phantom{-}.037 & \phantom{-}.160 & \phantom{-}.054 & .128 & \phantom{-}.037 \\
        - PCA + cluster & \phantom{-}.075 & \phantom{-}.075 & \phantom{-}.073 & \phantom{-}.067 & -.011 \\
        - SVD + cluster & \phantom{-}.036 & -.058 & \phantom{-}.020 & \phantom{-}.052 & \phantom{-}.017\\
		\bottomrule
	\end{tabular}
	\captionof{table}{Spearman $\rho$ obtained on KS2013-CoNLL. } \label{ks2013conll_table}
\end{figure}

\subsection{Ambiguity Analysis}

To investigate to what extent our models encode ambiguity at a word level, we turn to von Neumann entropy (VNE). 
For a density matrix $\rho = \sum_i p_i \vect{v_i}\vect{v_i}^{\top}$ the VNE is defined as
\begin{equation} \label{vne_def}
    S(\rho) = -\mathrm{tr}(\rho \ln{\rho}).
\end{equation}
This can be seen as an extension of Shannon entropy to matrices, and quantifies the amount of information encoded in a density matrix. 

We perform two analyses of ambiguity with VNE. First we test whether our density matrices model lexical ambiguity. We do this by investigating whether or not the measured ambiguity of a word's density matrix correlates with the number of meanings associated with the word. Secondly, we perform a systematic analysis of how ambiguity changes when words are composed into phrases. Using the four disambiguation data sets, we measure the VNE before and after composition, expecting ambiguity to decrease after composition. Something similar was done by \citet{piedeleu2015open}, at a smaller scale. 

For these experiments, we only report results for one variant of multi-sense Word2DM (cosine
similarity, 5 senses) and two variants of BERT2DM (SVD and PCA).

\begin{figure*}[t]
\centering
\includegraphics[width=\textwidth]{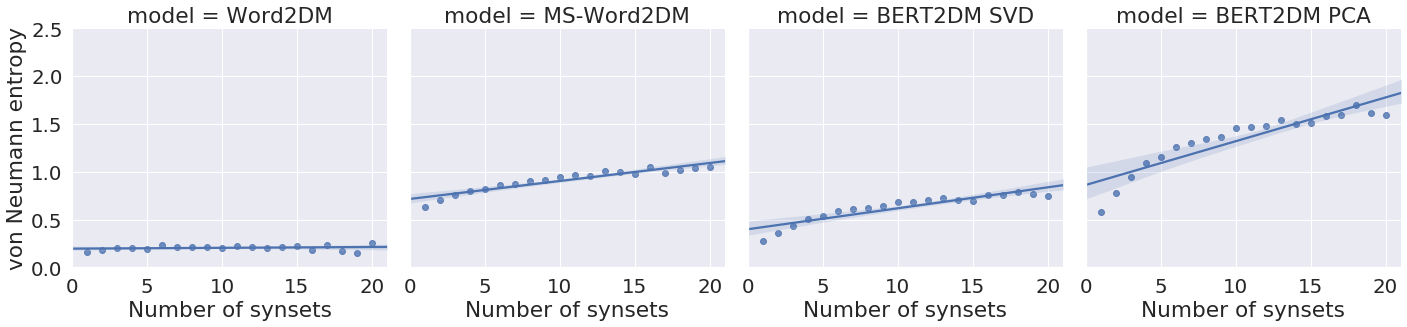}
\captionof{figure}{The average VNE of density matrices plotted against the number of WordNet synsets of words.} \label{ambiguity_plots}
\end{figure*}

\paragraph{Ambiguity and polysemy}
To determine the number of senses that a word has, we use WordNet synsets (senses) \citep{miller1995wordnet}. 
We compute the correlation  between the VNE of density matrices and the number of synsets associated with words. The correlation coefficients are shown in table \ref{ambiguity_coefficients} and the relationships are plotted in figure \ref{ambiguity_plots}. 

The results show that both BERT2DM and multi-sense Word2DM successfully encode how ambiguous words are. Word2DM exhibits a very low correlation and Context2DM (not plotted) shows none.

\begin{figure}[H]\small
    \centering
	\begin{tabular}{lcc}
		\toprule
	     Model & Pearson $r$ & Spearman $\rho$\\
	     \midrule
        Context2DM & -.081 & -.043\\
        Word2DM & \phantom{-}.053 & \phantom{-}.112\\
        MS-Word2DM & \phantom{-}.296 & \phantom{-}.295\\
        BERT2DM SVD & \phantom{-}.367 & \phantom{-}.418\\
        BERT2DM PCA & \phantom{-}\textbf{.405} & \phantom{-}\textbf{.463}\\
		\bottomrule
	\end{tabular}
	\captionof{table}{Correlation coefficient between the VNE of a word's density matrix and the number of WordNet synsets to which the word belongs.} 	\label{ambiguity_coefficients}
	\vspace{-0.5cm}
\end{figure}

\paragraph{Ambiguity and composition}
VNE allows us to measure how ambiguity evolves through composition. We can compare the ambiguity of a word to the ambiguity of a phrase containing the word. Seeing the context in which a word occurs can reveal which sense of the word is being employed, and should therefore reduce the amount of ambiguity present. 
The multi-sense Word2DM density matrix for the ambiguous word \emph{run} has a VNE of 1.491. After composition in the sentence \emph{The family run the hotel}, the sentence has a VNE of 0.144, so ambiguity has decreased.

We test whether this phenomenon holds true for our models on the disambiguation data sets, which consist of ambiguous verbs and disambiguating phrases. 
For each of the data sets we compute the average VNE of the verb density matrices. We compare this to the average VNE of the disambiguating phrases, where the density matrices are composed using different composition methods.  The results are presented in tables \ref{ml2008_comp_table} to \ref{ks2013_comp_table}.

In each of the tables, the leftmost column (Verb) contains the average von Neumann entropy of the ambiguous verbs alone, while the other columns contain the average von Neumann entropy after composing the disambiguating phrases with the different composition methods. Cases where the average von Neumann entropy decreases after composition (as expected) are highlighted in bold.

 The results are quite similar across the data sets.  Phaser emerges as the best method for decreasing ambiguity through composition. 
 This nicely supports the results of the disambiguation experiments, in which Phaser also emerged as the best composition method for disambiguating the meaning of ambiguous words through composition. Besides Phaser, none of the other composition methods reliably decrease the measured ambiguity.

\begin{figure}[t]		
    \small
	\begin{tabular}{@{}p{0.14\textwidth}p{\x\textwidth}p{\x\textwidth}p{\x\textwidth}p{\x\textwidth}p{\x\textwidth}}
		\toprule
	     & Verb & Mult & Add & Tensor & Phaser \\
		\toprule	
        Context2DM & 0.344 & 0.354 & 0.429 & 0.354 & \textbf{0.007}\\
        Word2DM & 0.103 & 0.257 & 0.707 & 0.257 & \textbf{0.019}\\
        MS-Word2DM & 0.953 & 1.751 & 1.537 & 1.751 & \textbf{0.468}\\
        BERT2DM SVD & 0.667 & \textbf{0.132} & 0.938 & \textbf{0.132} & \textbf{0.060}\\
		\bottomrule
	\end{tabular}
	\captionof{table}{Average VNE before and after composition on ML2008.} 
	\label{ml2008_comp_table}
\end{figure}

\begin{figure}[t]		
    \small
    %\centering ~%p{\x\textwidth}
	\begin{tabular}{@{}p{0.14\textwidth}p{\x\textwidth}p{\x\textwidth}p{\x\textwidth}p{\x\textwidth}p{\x\textwidth}}
		\toprule
	     & Verb & Mult & Add & Tensor & Phaser \\
		\toprule	
        Context2DM & 0.174 & 0.187 & 0.359 & 0.402 & \textbf{0.000}\\
        Word2DM & 0.245 & 0.522 & 1.096 & 0.797 & \textbf{0.038}\\
        MS-Word2DM & 1.444 & 1.821 & 1.906 & 4.012 & \textbf{0.224}\\
        BERT2DM SVD & 0.763 & \textbf{0.008} & 0.961 & \textbf{0.174} & \textbf{0.003}\\
		\bottomrule
	\end{tabular}
	\captionof{table}{Average VNE before and after composition on GS2011.} \label{gs2011_comp_table}
\end{figure}

\begin{figure}[t]		
    \small
	\begin{tabular}{@{}p{0.14\textwidth}p{\x\textwidth}p{\x\textwidth}p{\x\textwidth}p{\x\textwidth}p{\x\textwidth}}
		\toprule
	     & Verb & Mult & Add & Tensor & Phaser \\
		\toprule	
        Context2DM & 0.167 & \textbf{0.137} & 0.417 & 0.387 & \textbf{0.000}\\
        Word2DM & 0.140 & 0.623 & 1.445 & 0.908 & \textbf{0.005}\\
        MS-Word2DM & 1.340 & \textbf{1.181} & 2.043 & 3.534 & \textbf{0.029}\\
        BERT2DM SVD & 0.659 & \textbf{0.000} & 1.080 & \textbf{0.014} & \textbf{0.000}\\
		\bottomrule
	\end{tabular}
	\captionof{table}{Average VNE before and after composition on GS2012.} 
	\label{gs2012_comp_table}
\end{figure}

\begin{figure}[t]	
    \vspace{-0.2cm}
    \small
	\begin{tabular}{@{}p{0.14\textwidth}p{\x\textwidth}p{\x\textwidth}p{\x\textwidth}p{\x\textwidth}p{\x\textwidth}}
		\toprule
	     & Verb & Mult & Add & Tensor & Phaser \\
		\toprule	
        Context2DM & 0.233 & \textbf{0.183} & 0.524 & 0.537 & \textbf{0.000}\\
        Word2DM & 0.161 & 0.604 & 1.405 & 0.889 & \textbf{0.010}\\
        MS-Word2DM & 1.224 & 1.453 & 2.032 & 3.704 & \textbf{0.033}\\
        BERT2DM SVD & 0.689 & \textbf{0.000} & 1.124 & \textbf{0.026} & \textbf{0.000}\\
		\bottomrule
	\end{tabular}
	\captionof{table}{Average VNE before and after composition on KS2013-CoNLL.} 
	\label{ks2013_comp_table}
	\vspace{-0.25cm}
\end{figure}

\section{Conclusion and Future Work}

In this paper we addressed the problem of modelling ambiguity in NLP, and how ambiguous words can be disambiguated in context. We investigated density matrices as semantic representations for modelling ambiguity. Our results confirmed the value of density matrices over vector-based approaches. 
Equipped with a compositional framework, one of our density matrix models (multi-sense Word2DM) outperformed all other models (including existing compositional models and strong neural baselines) on most of the disambiguation tasks. We also performed a mathematical analysis of the ambiguity encoded by our models. This revealed that the density matrices built by two of our models (multi-sense Word2DM and BERT2DM) reflect true word level ambiguity. We have shown the value in designing neural models that learn density matrices from scratch. 

Possible directions for future work includes extending our models to larger datasets and longer sentences, and modifying techniques for differing sentences as in the Word in Context dataset \cite{pilehvar2019wic} and other WSD tasks. We focused on ambiguity here, but it would be possible to do similar experiments focused on other aspects of meaning such as metaphor or entailment, and examining how these interact with composition.

\section*{Acknowledgments}
Martha Lewis is supported by NWO Veni grant `Metaphorical Meanings for Artificial Agents'.

\bibliography{references}
\bibliographystyle{acl_natbib}

\clearpage
\appendix
\section{Word2DM Gradients} \label{app:word2dm}

The objective function that SGNS optimises at each prediction with regard to model parameters $\theta$ is
\begin{equation}\label{negative_sampling}
    J(\theta) = \log \sigma({v_t}^\top v_c) + \sum_{k=1}^K  \log \sigma(-{v_t}^\top v_{w_k})
\end{equation} 
where $v_t$ is the embedding of target word, $v_c$ is the embedding of the context word, and $v_1, v_2, ..., v_K$ are the embeddings of $K$ negative samples. By optimising equation \ref{negative_sampling} over a large corpus, skip-gram learns word embeddings that encode distributional information. 

Maximising equation \ref{negative_sampling} adjusts the embeddings of words occurring in the same context to be more similar and adjusts the embeddings of words that don't occur together to be less similar. 
This becomes clear when we consider the gradients used to update embeddings during training. We briefly recall the details of the gradient calculation so as to refer back to it later in this section. The derivative of equation \ref{negative_sampling} with respect to the target vector $v_t$ is
\begin{equation} \label{vt_gradient}
    \frac{\partial J}{\partial v_t} = (1 - \sigma({v_t}^\top v_c))v_c - \sum_{k=1}^K  (1 - \sigma({v_t}^\top v_{w_k}))v_{w_k}
\end{equation} 
which is used to update the target vector as follows:
\begin{equation} \label{vt_update}
v_t \leftarrow v_t + \alpha\frac{\partial J}{\partial v_t}
\end{equation}
The target vector is updated by adding the scaled context vector to it and subtracting the scaled negatively sampled vectors from it. The vectors are scaled proportionally to how dissimilar they are to the target vector. This ensures that the target vector is ``pulled closer'' to the true context vector and ``pushed away'' from the negative context vectors. It is this computationally simple training procedure which makes SGNS effective.

Word2DM extends SGNS to learn density matrices, replacing equation \ref{negative_sampling} with the following objective function:
\begin{equation}\label{word2dm_negative_sampling_appendix}
    J(\theta) = \log \sigma(\mathrm{tr}(A_t A_c)) + \sum_{k=1}^K  \log \sigma(-\mathrm{tr}(A_t A_{w_k})) 
\end{equation} 
where $A_t$ and $A_c$ are the density matrices of the target and context words respectively, $A_1, A_2, ..., A_K$ are the density matrices of $K$ negative samples, and $\theta$ is the set of weights of the intermediary matrices $B_t, B_c$ and $B_1, B_2, ..., B_K$. 

Computing this objective function requires multiple matrix multiplications. For each $\mathrm{tr}(A_t A_c)$ term (including the terms of the $K$ negative samples), the matrices $A_t$ and $A_c$ have to be computed respectively as $A_t = B_{t}B_{t}^\top$ and $A_c = B_{c}B_{c}^\top$ and then the matrix product $A_t A_c$ has to be computed. This means that, for each target-context prediction, we require $3(K+1)$ matrix multiplications. One of the most attractive features of SGNS is its computational efficiency, which enabled training on very large corpora in reasonable time. The introduction of multiple matrix multiplications into the objective function means that much of this efficiency is lost. In order to reduce the complexity of our model, we make use of the following property and lemma to find a new objective function that is computationally simpler, but equivalent to equation \ref{word2dm_negative_sampling_appendix}.

\begin{property}\label{property_trace_product}
The trace of the product of two matrices can be expressed as the sum of the element-wise products of their elements. If $A$ is an $n\times m$ matrix and $B$ is an $m \times n$ matrix, then the trace of the $n \times n$ matrix $AB$ can be computed as 
\[\mathrm{tr}(AB) = \sum_{i=1}^{n} \sum_{j=1}^{m} a_{ij} b_{ji}\] 
\end{property}

\begin{lemma}\label{lemma_word2dm_obj}
If $B_t$ and $B_c$ are $n \times m$ intermediary matrices, then trace of the matrix product $A_t A_c$ can be written as the sum of the squared elements of an $m \times m$ matrix $C = B_c^\top B_t$:
\[\mathrm{tr}(A_t A_c) = \sum_{i=1}^{m} \sum_{j=1}^{m} c_{ij}^2\]
\end{lemma}
\begin{proof}
We can express $\mathrm{tr}(A_t A_c)$ as a trace computation involving intermediary matrices $B_t$ and $B_c$:
\begin{align*}
\mathrm{tr}(A_t A_c) &= \mathrm{tr}(B_t B_t^\top B_c B_c^\top)
\end{align*}
Then we can use the cyclic property of the trace function to rewrite this as the product of a matrix $C$ and its transpose:
\begin{align*}
\mathrm{tr}(A_t A_c) &= \mathrm{tr}( B_c^\top B_t B_t^\top B_c) \\
                    &= \mathrm{tr}( B_c^\top B_t (B_c^\top B_t)^\top) \\
                    &= \mathrm{tr}( C C^\top), \quad \mathrm{where} \,\,\,  C = B_c^\top B_t 
\end{align*}
Now we can use property \ref{property_trace_product} to express this as the element-wise products of the elements of $C$ and its transpose:
\begin{align*}
\mathrm{tr}(A_t A_c) &= \sum_{i=1}^{m} \sum_{j=1}^{m} [C]_{ij} [C^\top]_{ji} \\
 &= \sum_{i=1}^{m} \sum_{j=1}^{m} c_{ij} c_{ij} \\
 &= \sum_{i=1}^{m} \sum_{j=1}^{m} c_{ij}^2 
\end{align*}
\end{proof}
This allows us to rewrite equation \ref{word2dm_negative_sampling_appendix} to find an equivalent objective function that requires fewer computations than straightforward matrix multiplication would. The objective function at each target-context prediction becomes
\begin{align}\label{word2dm_negative_sampling_simpler}
    J(\theta) =& \log \sigma(\sum_{i=1}^{m} \sum_{j=1}^{m}[B_c^\top B_t]_{ij}^2) \\
                &+ \sum_{k=1}^K  \log \sigma(-\sum_{i=1}^{m} \sum_{j=1}^{m}[B_{w_k}^\top B_t]^2_{ij}) . \nonumber
\end{align} 
By using the result of lemma \ref{lemma_word2dm_obj} we have reduced the number of matrix multiplications required for each target-context prediction from $3(K+1)$ to $(K+1)$. Density matrices are trained by maximising equation \ref{word2dm_negative_sampling_simpler} with respect to the intermediary matrices $B_t, B_c, B_{w_1}, ..., B_{w_K}$ over a large corpus. 

The model is trained using stochastic gradient descent. We now derive the gradients used to update $B_t$ during training, and subsequently show that these gradients lead to suboptimal updates to the density matrices during training. Deriving the gradient with respect to $B_c$ and $B_{w_k}$ would proceed similarly. To compute the gradients of equation \ref{word2dm_negative_sampling_simpler} we first rewrite it in terms of the elements of the $n \times m$ matrices $B_t$, $B_c$, and $B_{w_k}$: 
\begin{align}
    J(\theta) =& \log \sigma(\sum_{i=1}^{m} \sum_{j=1}^{m} (\sum_{l=1}^{n}b_{li}^{c} b_{lj}^{t} )^2) \\
    &+ \sum_{k=1}^K  \log \sigma(-\sum_{i=1}^{m} \sum_{j=1}^{m} (\sum_{l=1}^{n} b_{li}^{w_k} b_{lj}^{t})^2), \nonumber
\end{align} 
where $b^x_{pq}$ denotes the $pq$th element of $B_x$. We derive the gradient of this objective function with respect to $b^t_{pq}$, an element of the intermediary target word matrix $B_t$. In order to use the chain rule in gradient calculations we rewrite $J(\theta)$ as a composite function: 
\begin{equation}
    J(\theta) = \log \sigma(y(\theta)) + \sum_{k=1}^K  \log \sigma(z_k(\theta)),
\end{equation}
where 
\[\quad\quad y(\theta) =  \sum_{i=1}^{m} \sum_{j=1}^{m} (\sum_{l=1}^{n}b_{li}^{c} b_{lj}^{t} )^2 \quad\quad \mathrm{and} \]
\[z_k(\theta) = -\sum_{i=1}^{m} \sum_{j=1}^{m} (\sum_{l=1}^{n} b_{li}^{w_k} b_{lj}^{t})^2.\]
The derivative of $J$ with respect to $b^t_{pq}$ can now be computed as follows:%
{\small
\begin{align*}
\frac{\partial J}{\partial b_{pq}^t} =& \frac{\partial \log}{\partial \sigma} \frac{\partial \sigma}{\partial y} \frac{\partial y}{\partial b_{pq}^t} + \sum_{k=1}^K \frac{\partial \log}{\partial \sigma} \frac{\partial \sigma}{\partial z_k} \frac{\partial z_k}{\partial b_{pq}^t}\\
=&  \frac{1}{\sigma(y)}(1-\sigma(y))\sigma(y)\frac{\partial y}{\partial b_{pq}^t} \\
&+ \sum_{k=1}^K\frac{1}{\sigma(z_k)}(1-\sigma(z_k))\sigma(z_k)\frac{\partial z_k}{\partial b_{pq}^t}\\
=& (1-\sigma(y))\frac{\partial y}{\partial b_{pq}^t} + \sum_{k=1}^K(1-\sigma(z_k))\frac{\partial z_k}{\partial b_{pq}^t}\\
=& (1-\sigma(y)) \sum_{i=1}^m2b_{pi}^c \sum_{l=1}^n b_{li}^c b_{lq}^t \\
&- \sum_{k=1}^K(1-\sigma(z_k))\sum_{i=1}^m2b_{pi}^{w_k} \sum_{l=1}^n b_{li}^{w_k} b_{lq}^t\\
=& (1-\sigma(y)) 2[B_c B_c^\top B_t]_{pq} \\
&- \sum_{k=1}^K(1-\sigma(z_k))2[B_{w_k} B_{w_k}^\top B_t]_{pq} 
\end{align*}
}%
The last line in the above derivation is obtained by rewriting the summation expressions as equivalent matrix multiplications. We can now write the derivative of $J$ with respect to the full intermediary matrix $B_t$:
\begin{align} \label{Bt_gradient}
    \frac{\partial J}{\partial B_t} =& \, (1-\sigma(y(\theta))) 2 B_c B_c^\top B_t \\
    &+ \sum_{k=1}^K  (1- \sigma(z_k(\theta))) 2 B_{w_k} B_{w_k}^\top B_t  \nonumber
\end{align} 
As opposed to the gradients of Word2Vec (equation \ref{vt_gradient}), the gradients of Word2DM do not lead to simple and easily interpretable training updates. As discussed in the paragraph following equation \ref{vt_update}, in Word2Vec the target vector is made more similar to the context vector and less similar to the negative context vectors. Ideally we would like something similar to occur in Word2DM with density matrices, but equation \ref{Bt_gradient} shows that we lose the intuitive training updates of Word2Vec through the introduction of intermediary matrices. Furthermore, we can show that the gradients of Word2DM sometimes lead to unwanted consequences in training.

Consider the case where the density matrices of a target and context word are highly dissimilar. Recall from equation \ref{word2dm_negative_sampling_appendix} that the $y$ is in equation \ref{Bt_gradient} is the trace inner product of the density matrices $A_t$ and $A_c$ (the measure we use to quantify semantic similarity). The minimum value of the trace inner product of two density matrices is zero (this follows from the fact that density matrices are positive semi-definite), so two density matrices are highly dissimilar when their trace inner product is close to zero i.e. $y \approx 0$. From equation \ref{word2dm_negative_sampling_simpler} we can recall how $y$ can be written in terms of the intermediary matrices:
\[y = \sum_{i=1}^{m} \sum_{j=1}^{m}[B_c^\top B_t]_{ij}^2\]
Consider that $y \approx 0$ if and only if the elements of $B_c^\top B_t$ are close to zero in value, since squaring the elements in the summation makes them all positive. We have established the following equivalence:
\[
\mathrm{tr(A_t A_c)} \approx 0 \iff B_c^\top B_t \approx \mathrm{O},
\]
where $\mathrm{O}$ is the $m \times m$ matrix with all zero entries.
Consider how this will affect the target-context update during training. The first term of the gradient in equation \ref{Bt_gradient} becomes
\[
    (1-\sigma(y(\theta))) 2 B_c B_c^\top B_t = (1 - \sigma(0)) 2 B_c \mathrm{O} \approx \mathrm{O}
\]
so the target-context update becomes ineffective for true contexts. The update should make the density matrix of the target word more similar to that of the context word, but the gradient is so small that it makes this impossible. Moreover, the more dissimilar the target and context density matrices are before the update, the less effective the update will be. This is the opposite of the intended effect (achieved by Word2Vec) in which the magnitude of the target-context update should increase if the target and context representations are less similar. This is an example of how the introduction of intermediary matrices in Word2DM leads to suboptimal training updates. We ensure that our density matrices are positive semi-definite, but lose the guarantee that the algorithm will learn high-quality semantic representations.

\section{Hyperparameters for Word2DM and Multi-Sense Word2DM}

\paragraph{Word2DM additional details} We use a dynamic window size i.e. the size of each context window is sampled uniformly between 1 and the maximum window size. We also discard words that occur less than some minimum threshold and sub-sample frequently occurring words. Negative samples are drawn from a unigram distribution raised to the power of $\frac{3}{4}$. Furthermore, we train two density matrices for each word - one that represents it as a target word and another that represents it as a context word. 
After training we use the target density matrices as our final density matrices.  

\paragraph{Hyperparameters} We train our Word2DM and multi-sense Word2DM models on the ukWaC+Wackypedia corpus, consisting of 2.8 billion words. We use a window size of 5, a minimum word count of 50, 5 negative samples per positive context, and a subsampling rate of 1e-5. We train the model for 4 iterations of the ukWaC+Wackypedia corpus, using the Adam optimisation algorithm, a learning rate of 0.001, and 16 sentences per batch.

\end{document}